\ifdefined\XeTeXversion\else\ifdefined\pdfoutput\pdfoutput=1\fi\fi

\documentclass[10pt,twocolumn]{article}

\usepackage[preprint]{antintlpaper}

\usepackage[utf8]{inputenc}
\usepackage{inconsolata}
\usepackage{xspace}
\usepackage{float}
\usepackage{placeins}
\usepackage{tikz}
\usepackage{pifont}
\usetikzlibrary{arrows.meta,positioning,fit,calc}


\newcommand{\name}{\textsc{BenchCompass}\xspace}
\newcolumntype{Y}{>{\raggedright\arraybackslash}X}
\newcolumntype{P}[1]{>{\raggedright\arraybackslash}p{#1}}
\newcolumntype{C}[1]{>{\centering\arraybackslash}p{#1}}
\definecolor{ivblue}{HTML}{245B73}
\definecolor{ivteal}{HTML}{497D78}
\definecolor{ivgold}{HTML}{B6862C}
\definecolor{ivgray}{HTML}{F3F5F6}
\definecolor{ivline}{HTML}{9BA7AE}

\newcommand{\passsym}{\ding{51}}
\newcommand{\failsym}{\ding{55}}
\newcommand{\AntCorrespondingAuthor}{%
  \texorpdfstring{\textsuperscript{\ensuremath{\dagger}}}{ (corresponding)}}

\AntTitle{\name: From Scores to Signals for Training and Harness Decisions in Payment-Domain LLMs}
\AntRunningTitle{BenchCompass: Signals for Payment-Domain LLMs}
\AntAuthors{Sijie Dong\AntEqualContributor \and
  Wei Ren\AntEqualContributor \and
  Xuanwei Hu \and
  Jiawei Luo \and
  Zifan Wang \and
  Xiaoyun Feng \and
  Hui Cai \and
  Lyuxin Xue \and
  Peng Lu \and
  Jianshe Li \and
  Xin Zhang\AntCorrespondingAuthor \and
  Wei Wu\AntCorrespondingAuthor}
\AntAffiliations{Ant International}
\AntContact{Corresponding authors:
  \href{mailto:wuwei19850318@gmail.com}{wuwei19850318@gmail.com} and
  \href{mailto:evanzhangxin@gmail.com}{evanzhangxin@gmail.com}.\\
  \texttt{\{dongsijie.dsj,modi.rw\}@ant-intl.com}}

\AntFirstPageNote{%
  \parbox[b]{0.88\textwidth}{\raggedright
    \textsuperscript{*}\,Equal Contributions.
    \textsuperscript{\ensuremath{\dagger}}\,Corresponding Authors.\\[-1pt]
    }}
\AntDate{\today}
\AntAbstract{%
Payment operations are a critical financial infrastructure, but the value of large language models in this domain remains unclear because payment rules change quickly, evidence is fragmented, and decisions depend on transaction state, participant role, region, and payment rail.
Existing benchmarks do not isolate whether failures come from missing payment-rule knowledge, poor use of supplied evidence, or brittleness under imperfect harness inputs.
We introduce \name, a payment-domain benchmark whose construction pipeline builds scenario-grounded tasks from typed evidence packs, applies LLM-based quality checks, creates task-input attack variants, and reserves final item admission for domain experts.
The release contains an expert-reviewed Pro benchmark covering payment knowledge, context-grounded scenario reasoning, and Attacked Open robustness, plus a lower-assurance Normal pool for inspection and future curation.
Across 16 model variants, \name shows qualitatively different failure modes: missing parametric payment knowledge, incomplete reasoning over supplied rules, and failure to reject plausible but invalid workflows.
The benchmark remains unsaturated: the best frontier model reaches $89.6\%$ on Open Context-Grounded Reasoning and $81.7\%$ under attacked inputs, while a representative 32B open-weight model reaches $69.8\%$ and $42.6\%$. Benchmark data and code are available at
    \nolinkurl{https://github.com/ant-intl/BenchCompass}.
}

\begin{document}
\makeanttitle

\section{Introduction}
Large language models (LLMs) have demonstrated remarkable capabilities across a wide range of general and specialized evaluations, spanning multitask knowledge in MMLU \citep{hendrycks2021mmlu}, broad task coverage in BIG-bench \citep{srivastava2023beyond}, and expert-level scientific reasoning in GPQA \citep{rein2023gpqa}. Their capabilities have also been extensively evaluated in code generation benchmarks such as HumanEval \citep{chen2021evaluating}, continuously updated benchmarks such as LiveBench \citep{white2025livebench}, and long-context evaluations such as LongBench v2 \citep{bai-etal-2025-longbench}. Despite this broad evaluation landscape, payment remains comparatively underexplored. Payments constitute a foundational component of financial activity, yet the domain is neither a static body of knowledge nor a single homogeneous task. Payment systems operate as financial infrastructures that coordinate participants, processes, rules, and settlement obligations. Their behavior varies across payment products, jurisdictions, processing rails, and regulatory regimes, while correct decisions often depend on transaction state, timing, participant roles, and rail-specific operational constraints. Existing financial benchmarks evaluate broad financial capabilities \citep{xie2024finben}, filing-grounded question answering \citep{islam2023financebench}, and numerical or table-based financial reasoning \citep{chen-etal-2021-finqa,zhu-etal-2021-tat,chen-etal-2022-convfinqa}; however, they provide limited coverage of the fragmented, rule-intensive, and state-dependent reasoning required in real-world payment systems.

The complexity of payment systems and payment behaviors pose distinctive challenges for state-of-the-art LLMs, making their capability boundaries in the payment domain an important yet underexplored question for the AI community. A systematic assessment of current LLMs can benefit both AI researchers and payment practitioners. For AI researchers, it can reveal capability gaps and failure modes that remain insufficiently addressed, helping prioritize future efforts in model training and adaptation. For payment practitioners, it can provide a more comprehensive and reliable understanding of where current LLMs can—and cannot—support real-world payment workflows, thereby informing their appropriate and effective adoption. 

In this work, we investigate whether advanced LLMs are also capable of handling payment-domain tasks that require specialized domain knowledge, evidence-grounded reasoning, and reliable performance under imperfect payment-knowledge harness engineering. To this end, we begin by constructing a payment-domain benchmark, a process that presents four key challenges. First, payment knowledge evolves frequently, making benchmarks based on one-off expert authoring difficult to maintain. Second, valid payment questions are difficult to formulate because the correct answer often depends jointly on the transaction state, participant role, payment rail, jurisdiction, timing, and applicable rule hierarchy, while the supporting evidence is distributed across fragmented documents. Third, although LLMs can scale candidate generation, as demonstrated by recent efforts in automated benchmark construction \citep{li2025autobencher,li2024arena,bai2023lmexaminer}, they can also generate ambiguous, unsupported, template-like, or knowledge-free questions that are inadequate for rigorous payment-domain assessment. Finally, constructing adversarial cases and performing automatic scoring can introduce artificial difficulty: perturbations may inadvertently alter the intended reasoning or answer path \citep{yoran2024robust,hong-etal-2024-gullible,niu-etal-2024-ragtruth}, while LLM-based judges can introduce additional evaluation artifacts in the absence of stringent quality controls \citep{zheng2023judging,liu-etal-2023-g,liu-etal-2024-ecbd,gevers-etal-2025-benchmarks,zheng2025cheating,feuer2025judgment}. 

To address these challenges, we introduce \name, a domain-document-driven pipeline for constructing diagnostic benchmarks, instantiated in the payment domain. First, to reduce the maintenance burden imposed by rapidly evolving payment rules, \name uses an LLM-assisted pipeline to generate source-grounded candidate cases from versioned payment documents, replacing much of the repetitive manual authoring effort while retaining domain experts in the final admission loop. Second, to achieve both breadth and depth, \name combines typed evidence packs with 60 predefined payment scenarios and Bloom-level difficulty targets, enabling candidate tasks to span diverse payment settings while requiring controlled levels of rule understanding, application, analysis, and evaluation. Third, to mitigate errors introduced by LLM-based generation, \name applies a suite of LLM-based quality gates that assess evidence support, answerability, ambiguity, difficulty, and diagnostic validity. Fourth, to ensure that robustness cases remain answer-preserving rather than merely misleading, a separate task-input attack stage generates perturbed variants and rejects those that alter the intended reasoning or answer path. Finally, to further limit artifacts from both benchmark construction and automated judging, these automatic gates serve only as a triage mechanism, while final admission to the Pro benchmark is determined by human experts. 

Using \name, we construct a domain-expert-reviewed payment Pro benchmark comprising 306 cases: 95 multiple-choice questions that assess payment-domain knowledge, 96 open-ended questions that require context-grounded reasoning over complex payment scenarios, and 115 task-input attack cases that evaluate robustness under answer-preserving perturbations. Our evaluation shows that failures in the payment domain do not stem from a single capability gap. Comparisons between Closed and Open settings indicate that supplying relevant rules can resolve many knowledge-related failures, yet performance on clean open-ended tasks remains far from saturated: the best frontier model achieves $89.6\%$, whereas a representative 32B open-weight model reaches $69.8\%$. On the separately constructed attacked-input set, the corresponding scores drop to $81.7\%$ and $42.6\%$, respectively, revealing that models may have access to the correct rule yet still follow an invalid but operationally plausible reasoning path. These results establish \name as a diagnostic benchmark that distinguishes whether further improvement requires stronger domain knowledge, better context-grounded reasoning, or more reliable verification of rule applicability. We additionally retain a separately labeled, lower-assurance Normal pool for inspection and future curation, while all headline empirical results are reported exclusively on the expert-admitted Pro benchmark.

\name makes three contributions. First, we release a domain-expert-reviewed payment-domain Pro benchmark spanning domain knowledge, context-grounded scenario reasoning, and robustness to task-input attacks. Second, we present an auditable, document-driven construction pipeline that integrates typed evidence packs, predefined scenarios, Bloom-level difficulty controls, LLM-based quality gates, answer-preserving task-input attacks, and final expert review to produce broad-coverage, traceable benchmark cases. Third, we formulate benchmark outcomes as diagnostic model--case signals rather than aggregate scores alone, enabling failures to be attributed to missing domain knowledge, ineffective use of supplied payment rules, or brittleness under imperfect harness inputs.

\section{Related Work}

\paragraph{Domain benchmarks for LLMs.}
General benchmarks measure broad knowledge, reasoning, and freshness \citep{hendrycks2021mmlu,srivastava2023beyond,liang2023helm,rein2023gpqa,white2025livebench}. 
Domain benchmarks move evaluation into specialized settings, including legal reasoning \citep{guha2023legalbench}, broad finance \citep{xie2024finben}, filing-grounded QA \citep{islam2023financebench}, and financial reasoning over heterogeneous evidence \citep{chen-etal-2021-finqa,zhu-etal-2021-tat,chen-etal-2022-convfinqa}. 
These resources motivate specialized evaluation, but they do not target payment operations as a multi-party, stateful, rapidly changing infrastructure, nor do they separate payment knowledge, evidence-grounded scenario solving, and attack-input reliability.

\paragraph{Automatic benchmark construction.}
Recent systems use LLMs to scale benchmark construction: AutoBencher optimizes generated datasets for difficulty, novelty, and separability \citep{li2025autobencher}; BenchBuilder curates hard prompts from interaction logs \citep{li2024arena}; and Language-Model-as-an-Examiner generates and scores questions without fixed references \citep{bai2023lmexaminer}. Bloom-oriented work further argues that benchmark design should control cognitive demand, not only topic coverage \citep{huber-niklaus-2025-llms}. \name differs by constructing source-grounded diagnostic cases for a changing private domain: evidence packs, scenarios, difficulty targets, and task-input attack families specify what payment knowledge, workflow condition, reasoning demand, and harness failure mode a case tests, with domain experts deciding Pro admission.

\paragraph{Context use and deployment diagnosis.}
Retrieval and self-reflective retrieval provide external evidence \citep{lewis2020rag,asai2023selfrag}, but models can miss, underuse, or over-rely on supplied context \citep{liu-etal-2024-lost,bai-etal-2025-longbench,yang-etal-2025-100,goldman-etal-2024-really}. 
RAG robustness studies further show that irrelevant, adversarial, conflicting, or unsupported retrieved content can degrade answers \citep{yoran2024robust,hong-etal-2024-gullible,park-lee-2024-toward,shen-etal-2024-assessing,niu-etal-2024-ragtruth}. 
\name is related to this literature but does not evaluate retrieval quality or claim empirical superiority over RAG-robustness benchmarks. Its evaluation target is payment-domain model behavior under fixed source availability. In particular, Attacked Open changes payment task inputs through rule-applicability modes such as sub-rule collapse, rule coverage gap, phantom state, and prior inversion, rather than simply adding distractor passages.

\section{Task Formulation and Diagnostic Protocol}
\label{sec:protocol}

\name evaluates a model by comparing its behavior under controlled information conditions. The goal is not only to rank models, but to separate three practical questions: whether a model already has the relevant payment knowledge, whether it can use supplied payment evidence, and whether it remains reliable when the surrounding harness provides imperfect context. 

\subsection{Evaluation Settings}



We consider three evaluation settings. \emph{Closed} provides the model with only the question, testing whether it can answer domain-specific questions using knowledge encoded in its parameters. \emph{Open} additionally provides sufficient payment-domain evidence, evaluating whether the model can correctly interpret and apply the supplied evidence to solve the task. \emph{Attacked Open} is a separate open-book robustness setting in which the question and supplied context may contain irrelevant, stale, conflicting, poorly ordered, or near-miss information. A Task-Input Attacked item is considered valid only if the perturbed question and context continue to support the original reference answer.

Here, ``harness'' refers to the input-side and verification components surrounding a single model call, including context selection, context packing and ordering, prompt framing, and rule-applicability or verifier checks. \name does not evaluate an end-to-end retrieval system, tool-using agent, or multi-agent decomposition pipeline. Instead, \emph{Attacked Open} operationalizes harness brittleness through answer-preserving perturbations to the question and supplied context.

\subsection{Diagnostic Signals}

Diagnostic signals are assigned to \emph{model--case pairs}. A question is therefore not intrinsically ``known'' or ``hard'': the same case can expose different failures for different models. For base tasks, \name compares Closed and Open outcomes. For attacked-input tasks, \name reports Task-Input Attack robustness separately, because this setting tests behavior under an imperfect harness context rather than baseline knowledge access.

\begin{table*}[ht]
\centering
\caption{\textbf{Diagnostic signals.} Base signals come from Closed/Open transitions. Task-Input Attack signals are reported separately for answer-preserving attacked items. \passsym{} indicates pass and \failsym{} indicates fail.}
\label{tab:diagnosis}
\scriptsize
\setlength{\tabcolsep}{3pt}
\renewcommand{\arraystretch}{1.03}
\begin{tabularx}{\textwidth}{@{}P{0.22\textwidth}C{0.06\textwidth}C{0.08\textwidth}P{0.34\textwidth}Y@{}}
\toprule
\textbf{Signal} & \textbf{Closed} & \textbf{Open} & \textbf{Interpretation} & \textbf{Repair Hypotheses} \\
\midrule
\multicolumn{5}{@{}l}{\textit{Closed/Open base diagnosis}} \\
\textsc{Parametric Known} & \passsym & \passsym
& Model parameters contain the knowledge.
& Monitor drift. \\
\textsc{Context Rescued} & \failsym & \passsym
& Supplying context changes a Closed failure into an Open pass.
& Improve retrieval/context; consider CPT/SFT. \\
\textsc{Context-Use Gap} & \failsym & \failsym
& The model fails despite sufficient context; its context reasoning is incomplete.
& Improve reasoning training, verifier, or rubric. \\
\textsc{Context Interference} & \passsym & \failsym
& Context hurts an otherwise correct answer.
& Audit prompt, context packaging, and judge. \\
\midrule
\multicolumn{5}{@{}l}{\textit{Task-Input Attack diagnosis}} \\
\textsc{Perturbation Robust} & -- & TIA \passsym
& Correct under answer-preserving perturbation.
& Keep as a robustness control. \\
\textsc{Perturbation-Brittle} & -- & TIA \failsym
& Fails under answer-preserving perturbation.
& Improve harness. \\
\bottomrule
\end{tabularx}
\end{table*}

The repair hypotheses in Table~\ref{tab:diagnosis} are hypotheses rather than validated causal interventions. 
They connect the observed signals to intervention families studied in prior work, including domain/task-adaptive training \citep{gururangan-etal-2020-dont}, retrieval and context engineering \citep{lewis2020rag,asai2023selfrag}, and judge or verifier design \citep{zheng2023judging,feuer2025judgment}. Matched interventions for every diagnostic label are left as future work.

\section{Construction Pipeline}
\label{sec:pipeline}

\subsection{Pipeline Overview}

\begin{figure*}[t]
\centering
\includegraphics[width=0.9\textwidth]{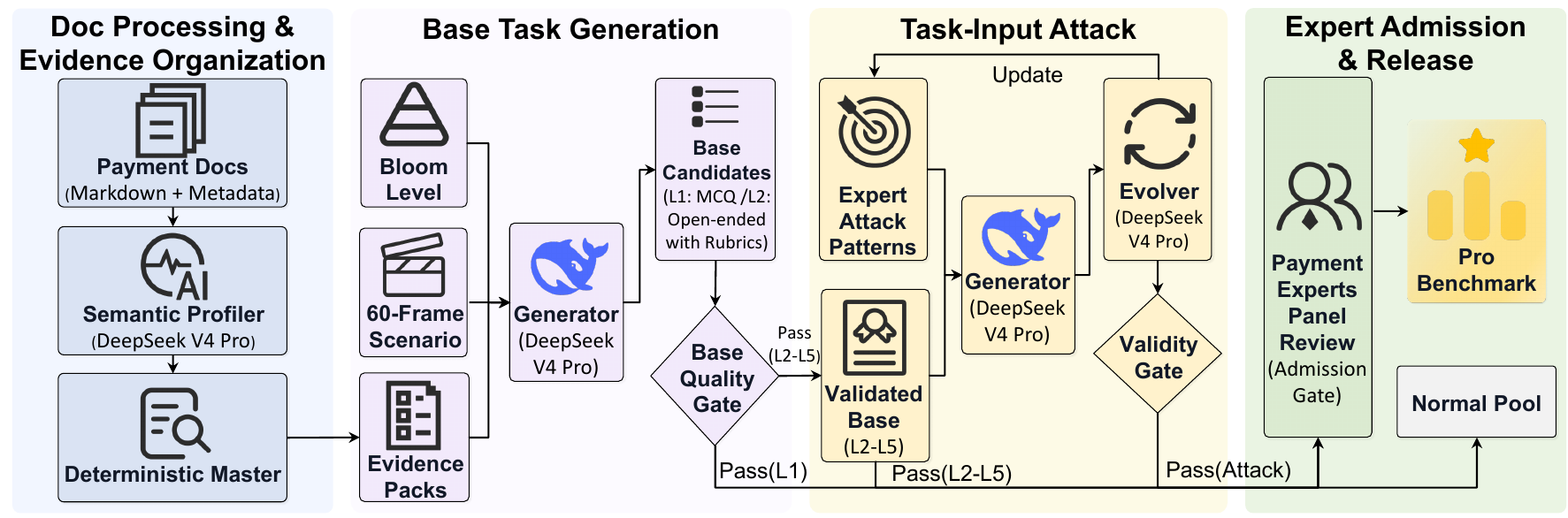}
\caption{\textbf{\name uses scenario-first evidence-pack construction.} The system builds structure-aware chunks, semantic profiles, typed evidence packs, and grounded scenarios before generating questions. }
\label{fig:pipeline}
\end{figure*}

Figure~\ref{fig:pipeline} summarizes \name. 
It turns a versioned payment-domain snapshot into structure-aware chunks, builds semantic profiles and evidence packs, and combines each pack with a scenario frame and Bloom difficulty level to generate base Closed/Open candidates. 
A base quality gate then checks ambiguity, answerability, and item quality before any attack is applied. 
Pattern-guided task-input attacks create Attacked Open candidates from validated answer paths, and attack-validity gates plus domain-expert admission decide whether each case enters Pro or a rejection pool. 

\subsection{Document Processing and Evidence Organization}

\name represents the versioned payment-domain snapshot as structure-aware Markdown chunks with audit metadata, including source category, heading path, source location, and rule or section identifiers. 
The source categories cover wallet, acquirer, clearing system, card network, regulator, platform, and related payment documents; Appendix Table~\ref{tab:source_role_distribution} reports the final admitted distribution. 
Before generation, DeepSeek V4 Pro builds a semantic profile for each chunk, summarizing its testable rule content, supported Bloom levels, and plausible payment scenarios. 
The pipeline then constructs evidence packs deterministically by linking each primary chunk to supporting chunks through inspectable relations such as section adjacency, shared rule identifiers, document family, payment system, currency, and profile role. 
Each evidence pack records the primary evidence, supporting evidence, and relation types, so reviewers can audit the source path behind every candidate task.

\subsection{Scenario-First Base Task Generation}

Base construction is scenario-first: each candidate is generated by pairing an evidence pack with an operational payment scenario and a target Bloom level. 
We use Bloom's revised taxonomy as the difficulty axis \citep{anderson2001taxonomy,huber-niklaus-2025-llms}, covering L1--L5: L1 targets payment knowledge, while L2--L5 test understanding, application, analysis, and exception or precedence evaluation over supplied evidence. 
We exclude L6 because open-ended creation tasks are difficult to score with stable reference answers.

The evidence pack anchors source support, while the scenario frame turns the source material into an operational payment problem rather than a decontextualized rule lookup. 
Our scenario bank contains 60 deterministic frames, with five frames for each of 12 payment-source categories, covering settings such as settlement, compliance review, disputes, product design, policy interpretation, message migration, onboarding, treasury operations, and liquidity management.
DeepSeek V4 Pro then generates structured JSON candidates from the Bloom target, evidence pack, and scenario frame, with instructions to ground the question, answer, and rubric in the source text. 
We refer to L1 items as \emph{Payment Knowledge} cases and L2--L5 items as \emph{Context-Grounded Reasoning} cases.
Candidates missing required answers, source quotes, choices, or rubrics are rejected.
The generator uses level-specific templates rather than bare Bloom labels: L1 recall, L2 explanation, L3 rule application, L4 interacting-rule or workflow analysis, and L5 exception, conflict, or precedence resolution. Quality gates and expert admission check compliance.

\subsection{Task-Input Attack Generation}

Attacked Open tasks start from L2--L5 base items that are rescued by Open evaluation, so each seed item already has a source-supported answer path. 
The attack stage, therefore, tests whether a model still follows the evidence when the user-facing scenario becomes misleading or underspecified. 
For each seed, an attack selector chooses one or two Bloom-compatible patterns from an expert-initialized attack pool covering real-world failure scenarios; Appendix Table~\ref{tab:attack_patterns} lists both the initial and pipeline-evolved attack patterns. 
An LLM generator then rewrites the task input with answer-preserving perturbations, while retaining source quotes, a reference answer, and grading rubrics.

The process is iterative: validity checks and construction-time Open correctness are used only to select useful candidate variants, not as final evaluation scores. 
The pipeline updates attack-pattern hit rates and records newly emerged patterns across rounds. 
Perturbations that change the intended answer, remove the evidence path, or make the rubric ambiguous are rejected, and valid Attacked Open cases are evaluated separately from the Closed/Open diagnostic tasks.

\subsection{Quality Control, Review, and Admission}

Quality control combines general validity checks with payment-specific filters. 
LLM gates verify format, evidence sufficiency, answerability, ambiguity, rubric fairness, diagnostic value, and attacked-input validity. 
A domain-specificity gate rejects or rewrites candidates whose answer no longer depends on payment actors, states, rails, message fields, clearing or settlement concepts, or regulatory scope.

Automatic gates are used only for triage. 
Each finalist retains source evidence, quote maps, rubrics, model outputs, judge decisions, rejection reasons, and review metadata. 
Four payment-domain practitioners first jointly reviewed 30 calibration cases (10 per subset) to align a structured protocol covering question/rubric validity, judge correctness, source support, specificity, diagnostic value, duplication, and attack validity; this was protocol calibration, not full-pool inter-annotator agreement. The remaining candidates were partitioned into four disjoint subsets, each independently reviewed by a different practitioner, and then merged for joint adjudication. Open-book cases answered incorrectly by Gemini 3.1 Pro, GPT-5.5, or Claude Opus 4.7 received an additional two-expert validity review. Only cases with no unresolved issues were admitted; borderline, ambiguous, weakly diagnostic, duplicate, source-unstable, or defective cases were held out.
The Normal pool is marked with lower assurance and retained only for inspection, development, and future curation; it is excluded from headline Pro evaluation claims.

\section{Experiments}
\subsection{Experimental Setup}

We evaluate \name on the expert-admitted Pro benchmark, covering Payment Knowledge MCQ, Context-Grounded Reasoning QA, and Task-Input Attack QA; the lower-assurance Normal pool is excluded from headline results. 
We test frontier closed-source, large open-weight, and mid-scale open-weight models through Alibaba Bailian and OpenRouter, using deterministic pass@1 decoding. 
Base tasks are evaluated in Closed and Open settings, while Task-Input Attack tasks are evaluated by attack family.

Headline OpenQA scoring uses a self-deployed DeepSeek V4 Flash judge in non-thinking mode, served on four H200 instances with four H200 GPUs each. 
To test judge-family sensitivity, we conduct a separate calibrated re-scoring analysis on fixed OpenQA outputs from three non-DeepSeek answer models: Gemini 3.1 Pro Preview, GLM 5.1, and Gemma 4 26B-A4B IT. DeepSeek V4 Flash, Qwen3.5-397B-A17B, and Qwen3.6-27B receive identical questions, contexts, reference answers, rubrics, and model outputs. This sensitivity analysis does not modify the headline scores.
Because DeepSeek V4 Pro is used during construction and also evaluated as an answer model, we treat its score as a reference model result rather than evidence for benchmark validity. 
Pro admission is decided by quality gates and domain experts, and evaluated models see only the final question and source context.
Full model IDs, provider routes, thinking settings, prompts, decoding settings, and judge configuration appear in Appendix~\ref{app:model_settings}, Table~\ref{tab:model_settings_appendix}.

\subsection{Benchmark Composition}

\begin{table}[t]
\centering
\caption{\textbf{Pro benchmark composition.} Counts are expert-admitted cases in the final benchmark.}
\label{tab:dataset_overview}
\scriptsize
\setlength{\tabcolsep}{2.6pt}
\renewcommand{\arraystretch}{1.13}
\begin{tabularx}{\columnwidth}{@{}P{0.30\columnwidth}C{0.08\columnwidth}C{0.14\columnwidth}Y@{}}
\toprule
\textbf{Subset} & \textbf{N} & \textbf{Format} & \textbf{Evaluation target} \\
\midrule
Payment Knowledge & 95 & MCQ & Closed/Open; L1 payment-rule knowledge \\
Context-Grounded Reasoning & 96 & Open QA & Closed/Open; L2--L5 scenarios (38/29/19/10) \\
Task-Input Attack & 115 & Open QA & Open only; answer-preserving task-input attacks \\
\bottomrule
\end{tabularx}
\end{table}

Table~\ref{tab:dataset_overview} summarizes the payment-domain release produced by \name. The final Pro benchmark contains 306 cases admitted after expert review: 95 Bloom L1 Payment Knowledge cases, 96 Bloom L2--L5 Context-Grounded Reasoning cases, and 115 Task-Input Attack cases. 
Payment Knowledge cases test closed-book payment knowledge. 
Context-Grounded Reasoning cases are evaluated in both Closed and Open settings. Task-Input Attack cases are evaluated with context and test robustness to imperfect harness engineering.

\begin{figure}[t]
\centering
\includegraphics[width=0.85\columnwidth]{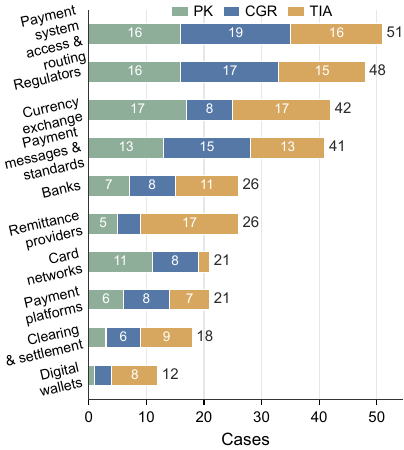}
\caption{\textbf{Payment-domain coverage across three benchmark conditions.} Each case is counted once by its primary evidence class. PK is L1 Payment Knowledge cases, CGR is L2-L5 Context-Grounded Reasoning cases, and TIA is Task-Input Attack cases.}
\label{fig:source_role_distribution}
\end{figure}

Figure~\ref{fig:source_role_distribution} reports payment-domain coverage. Each case is assigned to one primary evidence class, so the counts sum to the release size. The largest categories are payment-system access and routing, regulators, currency exchange, and payment-message standards. The remaining cases cover remittance providers, banks, payment platforms, card networks, clearing and settlement, and digital wallets. This distribution reflects the source material needed for payment operations rather than a uniform topic quota.
Source roles are not uniformly sampled because their operational scope and public-document availability differ. No class dominates: Payment-system access and routing accounts for 16.7\%, Regulators 15.7\%, Currency exchange 13.7\%, and Payment messages and standards 13.4\%. We therefore treat coverage as a documented release snapshot, not an exhaustive payment-domain population.
L5 contributes 10/96 CGR cases (10.4\%), providing exception/precedence coverage but not supporting standalone rankings or stable level-specific conclusions. Appendix Table~\ref{tab:app_dataset_stats} summarizes construction statistics.

\subsection{Main Results}

Table~\ref{tab:closed_open_main} reports the main diagnostic accuracies on the expert-admitted Pro benchmark. 
We avoid treating small differences between nearby models as fine-grained rankings. Instead, we focus on larger patterns: Closed-to-Open gains, lower accuracy on the separately constructed Attacked Open set, and robustness gaps across model groups.
Bootstrap standard errors and selected descriptive deltas are reported in Appendix Table~\ref{tab:app_bootstrap_se}.
For example, the three frontier models span only $14.6$--$36.5\%$ on CGR-CB but reach $82.3$--$89.6\%$ on CGR-OB; these large within-condition transitions, rather than nearby leaderboard positions, are the primary empirical comparison.
For these frontier models, bootstrap standard errors on CGR-CB/CGR-OB are $3.1$--$4.9$ percentage points, whereas their Closed-to-Open gains are $45.8$--$75.0$ points (Appendix Table~\ref{tab:app_bootstrap_se}); we therefore avoid fine-grained claims about nearby model scores.

\begin{table}[t]
\centering
\caption{\textbf{Diagnostic results} \textbf{PK-CB}: Payment Knowledge (Closed Book); \textbf{CGR-CB}: Context-Grounded Reasoning (Closed Book); \textbf{CGR-OB}: Context-Grounded Reasoning (Open Book); \textbf{TIA-OB}: Task-Input Attack (Open Book). Models are categorized by scale and access, sorted by Task-Input Attack (OB). Because each subset contains a finite number of expert-admitted cases, we interpret small point differences cautiously. \protect Payment Knowledge Open Book (PK-OB) is reported separately because it is saturated: 15/16 models score 95/95 ($100\%$), and Ministral 3 8B scores 93/95 ($97.9\%$).}
\label{tab:closed_open_main}
\scriptsize
\setlength{\tabcolsep}{4pt}
\renewcommand{\arraystretch}{1.06}
\begin{tabularx}{\columnwidth}{@{}Xcccc@{}}
\toprule
\textbf{Model} & \textbf{PK-CB} & \textbf{CGR-CB} & \textbf{CGR-OB} & \textbf{TIA-OB} \\
\midrule
\multicolumn{5}{@{}l}{\textbf{Closed-source API Models}} \\
\midrule
Gemini 3.1 Pro Preview & \textbf{80.0} & 14.6 & \textbf{89.6} & \textbf{81.7} \\
GPT-5.5 & 77.9 & \textbf{36.5} & 82.3 & 80.0 \\
Claude Opus 4.7 & 68.4 & 25.0 & 82.3 & 74.8 \\
\midrule
\multicolumn{5}{@{}l}{\textbf{Open-weight Large Models}} \\
\midrule
DeepSeek V4 Pro & 60.0 & 16.7 & 81.3 & 80.0 \\
Qwen3.5 397B-A17B & 62.1 & 14.6 & 84.4 & 72.2 \\
GLM 5.1 & 65.3 & 15.6 & 74.0 & 70.4 \\
Qwen3.5 122B-A10B & 59.0 & 9.4 & 84.4 & 70.4 \\
MiniMax M2.5 & 53.7 & 9.4 & 75.0 & 60.9 \\
Mistral Large 3 & 47.4 & 8.3 & 64.6 & 51.3 \\
Llama 4 Maverick & 54.7 & 8.3 & 68.8 & 46.1 \\
\midrule
\multicolumn{5}{@{}l}{\textbf{Open-weight Compact Models}} \\
\midrule
Gemma 4 31B IT & 57.9 & 9.4 & 82.3 & 74.8 \\
Gemma 4 26B-A4B IT & 61.1 & 7.3 & 82.3 & 60.9 \\
Qwen3 32B & 48.4 & 5.2 & 69.8 & 42.6 \\
Ministral 3 14B & 49.5 & 2.1 & 57.3 & 37.4 \\
Qwen3 8B & 44.2 & 1.0 & 58.3 & 36.5 \\
Ministral 3 8B & 51.6 & 2.1 & 50.0 & 29.6 \\
\bottomrule
\end{tabularx}
\end{table}

\subsubsection{Closed/Open Diagnostic Results}

The base evaluation separates accessible parametric knowledge from the ability to use the supplied payment context. 
The first key finding is that many payment tasks are knowledge-access failures rather than unsolvable reasoning failures. 
PK-CB ranges from $44.2\%$ to $80.0\%$, while open-book payment knowledge is nearly saturated and therefore omitted from the table. 
PK-CB ranges from $44.2\%$ to $80.0\%$, whereas PK-OB
ranges from $97.9\%$ to $100.0\%$: 15 models answer all 95 items
correctly, and Ministral 3 8B answers 93/95.
The gap is sharper on CGR: even GPT-5.5 reaches only $36.5\%$ in CGR-CB, and most models remain below $20\%$; with source context, CGR-OB rises to $50.0$--$89.6\%$. 
To assess wording sensitivity, we randomly audited 20 Gemini 3.1 Pro CGR-CB failures rescued in CGR-OB. In 19/20, the Closed answer omitted or misstated a substantive rule, condition, threshold, or workflow, or added an unsupported claim; only 1/20 was substantively correct but insufficiently aligned with the rubric's expected formulation. Thus, Closed-to-Open gains primarily correct source-specific errors, although formulation sensitivity contributes occasionally. Appendix Figure~\ref{fig:app_context_rescue_heatmap} reports rescue rates by source class, not direct estimates of parametric-knowledge absence.

The second key finding is that Open context still leaves a real context-use gap. No model reaches $90\%$ on CGR-OB, and the remaining errors are not mainly simple fact omissions. 
Across six models, manual audit identifies multi-rule synthesis and scope/exception boundaries as the leading error modes (26 model--case errors each), followed by over-expanded answer paths and modal/threshold imprecision (Appendix Table~\ref{tab:app_open_failure_modes}). Appendix Table~\ref{tab:app_diagnostic_cases} traces three cases from the missed rule to the triage implication, distinguishing context-rescued no-context failures from persistent source-grounded reasoning errors requiring better context use, verification, or adaptation.

As supplementary actionability checks on the same 95 Payment Knowledge items, a Codex corpus-access harness raised Qwen3.6-27B from 61.05\% to 92.63\%, while payment-domain QA SFT with 10\% general data raised it to 94.74\%. These checks support context access and training as actionable routes without causally validating every triage implication.

\subsubsection{Robustness against Task-Input Attacks}

\begin{figure*}[t]
\centering
\includegraphics[width=\textwidth]{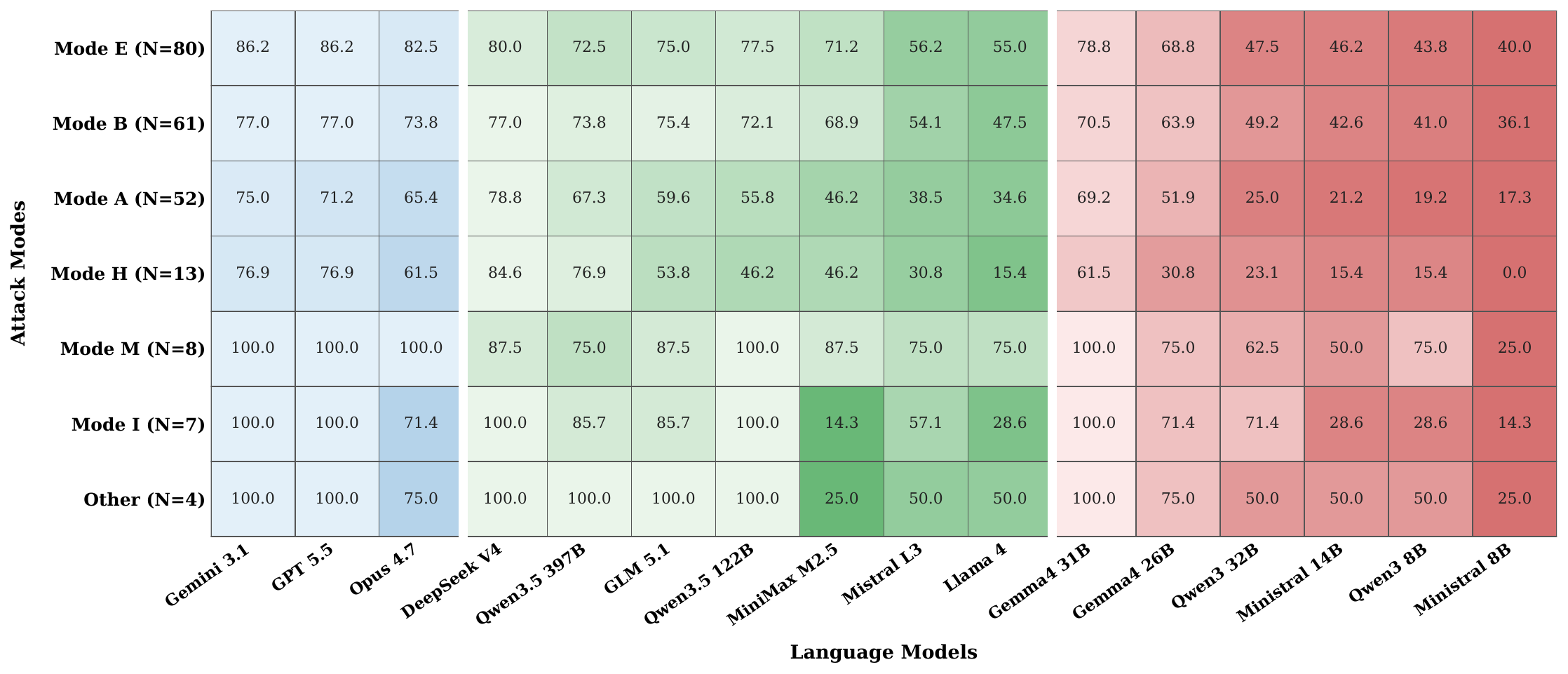}
\caption{\protect\textbf{Robustness by attack mode.} Cells report descriptive accuracy (\%) for cases carrying each attack-mode label. Labels overlap: 110 of the 115 cases carry two labels, and five carry one, yielding 225 mode mentions. A/B/E are the better-represented modes; H/M/I/D\_E/G(Other) are low-count and exploratory. Darker cells indicate lower accuracy. Attack definitions are provided in Appendix Table~\ref{tab:attack_patterns}.}
\label{fig:attack_modes_heatmap}
\end{figure*}

Task-Input Attack evaluation uses the fixed expert-admitted Attacked Open set. 
Each case remains answerable from the supplied context, but the user-facing scenario contains targeted perturbations that make the correct evidence path harder to follow. 
Thus, TIA-OB accuracies are final evaluation results, not construction-time attack success rates.

Table~\ref{tab:closed_open_main} reports clean CGR-OB accuracy alongside accuracy on the separately constructed TIA-OB set. 
Frontier models such as Gemini 3.1 Pro Preview and GPT-5.5 remain above $80\%$ on TIA-OB, but several otherwise strong models score lower on attacked inputs than on clean Open cases. 
For example, Qwen3.5 397B-A17B reaches $84.4\%$ on CGR-OB but $72.2\%$ on TIA-OB, and Llama 4 Maverick reaches $68.8\%$ versus $46.1\%$. 
This suggests that strong clean-context evidence use does not necessarily imply robustness to misleading task inputs, especially for lower-capacity models.

Figure~\ref{fig:attack_modes_heatmap} decomposes TIA-OB performance by attack-mode label. We focus on the three better-represented modes: Mode A (Sub-rule Premise Collapse; $N=52$) tests whether a model stops a downstream workflow when its parent prerequisite no longer holds; Mode B (Rule Coverage Gap Over-generalization; $N=61$) tests whether it avoids extending a rule to an unsupported edge case; and Mode E (Prior Inversion; $N=80$) tests whether it follows the supplied rule when that rule conflicts with a familiar operational prior. A-tagged cases have the lowest observed cross-model mean accuracy ($49.8\%$, versus $62.5\%$ for B and $66.7\%$ for E), and even frontier models reach only $65.4$--$75.0\%$ on A. This suggests that enforcing hierarchical rule applicability remains difficult even when the relevant rule is present. Because labels overlap and case mixtures differ, these results are descriptive rather than a causal ranking of attack difficulty. The low-count H/M/I/D\_E/G modes remain exploratory; their definitions and construction procedures are reported in Appendix Table~\ref{tab:attack_patterns}. Our headline robustness conclusion therefore relies on the complete 115-case TIA-OB result.

\subsection{Scaling and Thinking-Mode Effects}

Appendix Table~\ref{tab:app_scaling_thinking} shows that model size and thinking mode affect different diagnostic subsets differently. 
In the Qwen3 non-thinking sweep, L2--L5 Open accuracy rises from 53.1\% at 4B to 75.0\% at 235B, and Attacked Open accuracy rises from 28.7\% to 51.3\%, while L1 Closed remains within 50.5--58.9\%. 
For Gemma, thinking mode improves Attacked Open accuracy for both tested models, but its effect on ordinary Open accuracy is mixed.

\subsection{Construction Quality Analysis}
Appendix Table~\ref{tab:construction_funnel} summarizes the de-duplicated expert-review finalist funnel after automatic filtering and LLM-based triage. 
After automated quality filtering and deduplication, 506 candidate items entered expert review; four payment-domain experts admitted 306 of them to Pro (60.5\%). Appendix Table~\ref{tab:app_exclusion_reasons} summarizes the primary reasons for holding out the remaining candidates.
To assess admission reliability, 60 randomly sampled candidates (20 per subset) were blindly re-reviewed under the same criteria by an original practitioner who had not conducted that item's first-pass review. Binary agreement was 57/60 (95.0\%; Cohen's $\kappa=0.900$); the three disagreements involved borderline provider specificity or diagnostic value, not unsupported answers or incorrect rubrics. To check generator portability, we replaced DeepSeek V4 Pro with Qwen3.5-397B-A17B while keeping the downstream pipeline unchanged. Of 100 generated candidates, 62 were usable after downstream QC and conservative manual review. This shows that the code path can produce usable candidates with another strong model, but does not establish generator-model invariance.

\subsection{Judging Reliability Checks}

Four human reviewers audited 255 visible-verdict OpenQA judge decisions: all 115 Attacked Open decisions and 70 sampled decisions from each of Base Closed and Base Open. 
For open-book cases failed by frontier models, two reviewers cross-checked the case and judge decision; disagreement made the case ineligible for the benchmark.
As shown in Table~\ref{tab:human_scoring_audit}, reviewers accepted 251/255 decisions (98.4\%); all audited Base Open and Attacked Open decisions were accepted.


Under a shared calibrated protocol, Qwen3.5-397B and Qwen3.6-27B re-score 921 fixed OpenQA outputs from Gemini 3.1 Pro, GLM 5.1, and Gemma 4 26B-A4B. Their agreement with DeepSeek V4 Flash is 93.05\% and 91.97\%, with correctness-rate shifts of $-2.17$ and $-1.74$ points; per-model agreement exceeds 91\% (Appendix Table~\ref{tab:secondary_judge_agreement}). This separate sensitivity analysis reduces judge-family concerns without replacing headline scores and cautions against overinterpreting small differences.

For same-judge repeatability, three DeepSeek V4 Flash re-scorings of fixed GPT-5.5, GLM 5.1, and Qwen3 32B outputs across CGR-CB/CGR-OB/TIA-OB produce parseable JSON for all 2,763 calls and a maximum accuracy range of 2.08 points, well below the main Closed/Open transitions.

\begin{table}[t]
\centering
\caption{\textbf{Human audit of OpenQA judging.} Accept agrees with the judge, Partial marks a minor issue, and Wrong marks an incorrect judge decision; MCQ items are excluded.}
\label{tab:human_scoring_audit}
\scriptsize
\setlength{\tabcolsep}{2.4pt}
\renewcommand{\arraystretch}{1.05}
\begin{tabularx}{\columnwidth}{@{}P{0.25\columnwidth}*{4}{C{0.105\columnwidth}}C{0.22\columnwidth}@{}}
\toprule
\textbf{Evaluation} & \textbf{N} & \textbf{Accept} & \textbf{Partial} & \textbf{Wrong} & \textbf{Accept rate} \\
\midrule
Base Closed & 70 & 66 & 3 & 1 & 94.3 \\
Base Open & 70 & 70 & 0 & 0 & 100.0 \\
Attacked Open & 115 & 115 & 0 & 0 & 100.0 \\
\midrule
\textbf{All OpenQA} & \textbf{255} & \textbf{251} & \textbf{3} & \textbf{1} & \textbf{98.4} \\
\bottomrule
\end{tabularx}
\end{table}

\FloatBarrier

\section{Conclusion}
We introduced \name, a payment-domain benchmark and construction pipeline for evaluating LLMs on dynamic, document-grounded payment workflows. 
The benchmark shows that payment failures are not a single capability gap: models may lack domain knowledge, fail to compose supplied rules, or follow a plausible but invalid workflow despite having the relevant context. 
By separating these behaviors through Closed, Open, and Attacked Open settings, \name turns benchmark scores into diagnostic signals that can help prioritize future work on domain training, context use, and rule-applicability verification.

\section*{Limitations}

This paper studies payment-domain workflows rather than general finance. The Pro benchmark is expert-reviewed, but it is still a finite snapshot of public or release-reviewable payment documents, so scores should be interpreted relative to the released source snapshot rather than as permanent model properties.

\name evaluates use of supplied context, not end-to-end retrieval. The Closed, Open, and Attacked Open labels are behavioral diagnostic signals rather than causal explanations; matched repair experiments are needed to verify which interventions improve specific failure sets. Bloom levels and attack patterns are coverage devices, not a complete theory of payment reasoning. OpenQA scoring uses structured LLM judging with human audit and a secondary-judge check, but residual rubric sensitivity, judging errors, and generator/judge model-family overlap may remain.

\section*{Ethical Considerations}

\name is intended for benchmark evaluation and engineering diagnosis, not for payment, financial, legal, compliance, or customer-facing advice. The benchmark uses public or release-reviewed payment documents and does not use private customer records, transaction logs, account data, or personally identifiable information. Cases mentioning real institutions test the interpretation of public rules, not institutional behavior or risk.

The release includes runnable benchmark cases, evidence snippets, source provenance, reference answers, rubrics, prompts, model outputs, judge decisions, human audit labels, and run configuration. Full original documents are redistributed only when licensing permits. Attacked Open cases are answer-preserving robustness probes, not instructions for fraud, evasion, or bypassing controls. Human reviewers were informed of research use, reviewer identities are not released, and AI-assisted construction, coding, analysis, and editing were reviewed by the authors before release.

\bibliographystyle{plainnat}
\bibliography{custom}

\clearpage
\appendix

\begin{table*}[!t]
\section{Payment Source Coverage}
\medskip
\centering
\caption{\textbf{Payment source categories represented in the Pro benchmark.} Categories match Figure~\ref{fig:source_role_distribution}. Evidence documents are distinct source files within each category; the total row reports unique source files across the admitted Pro benchmark.}
\label{tab:source_role_distribution}
\footnotesize
\setlength{\tabcolsep}{4pt}
\renewcommand{\arraystretch}{1.06}
\begin{tabularx}{\textwidth}{@{}P{0.25\textwidth}rrrrrY@{}}
\toprule
\textbf{Source category} & \textbf{Ev. docs} & \textbf{L1} & \textbf{L2--L5} & \textbf{Attack} & \textbf{Total} & \textbf{Typical coverage} \\
\midrule
Payment system access \& routing & 11 & 16 & 19 & 16 & 51 & Participant access, routing, FPS/CHATS/RTGS-like systems, payment rails, and operational flow. \\
Regulators & 11 & 16 & 17 & 15 & 48 & Licensing, AML/CFT, consumer protection, data protection, and supervisory rules. \\
Currency exchange & 11 & 17 & 8 & 17 & 42 & FX conversion, exchange-rate disclosure, transfer limits, and cross-border currency constraints. \\
Payment messages \& standards & 15 & 13 & 15 & 13 & 41 & ISO 20022, SWIFT-like message standards, field requirements, and technical guidance. \\
Banks & 11 & 7 & 8 & 11 & 26 & Commercial and digital bank account terms, e-banking, onboarding, and account-state rules. \\
Remittance providers & 13 & 5 & 4 & 17 & 26 & Transfer flow, fees, delivery channels, arrival time, refunds, and service restrictions. \\
Card networks & 4 & 11 & 8 & 2 & 21 & Card-network rules, dispute handling, chargeback, tokenization, and merchant constraints. \\
Payment platforms & 6 & 6 & 8 & 7 & 21 & Cross-border collection, merchant settlement, PSP accounts, notifications, and platform risk controls. \\
Clearing \& settlement & 6 & 3 & 6 & 9 & 18 & Clearing-house rules, settlement timing, value date, reconciliation, and exception handling. \\
Digital wallets & 6 & 1 & 3 & 8 & 12 & Stored value, top-up, withdrawal, QR payment, KYC, and wallet transfer rules. \\
\midrule
\textbf{Total} & \textbf{73} & \textbf{95} & \textbf{96} & \textbf{115} & \textbf{306} & \textbf{Final domain-expert-reviewed Pro benchmark.} \\
\bottomrule
\end{tabularx}
\end{table*}


\begin{table*}[t]
\centering
\caption{\protect \textbf{Benchmark statistics and prompt targets.} Context is measured in characters. Source segments count the flattened `Source' blocks in the released context and summarize evidence-pack size. Rubric counts are means for OpenQA.}
\label{tab:app_dataset_stats}
\scriptsize
\setlength{\tabcolsep}{3.1pt}
\renewcommand{\arraystretch}{1.08}
{
\begin{tabularx}{\textwidth}{@{}P{0.17\textwidth}rP{0.17\textwidth}rrrY@{}}
\toprule
\textbf{Subset} & \textbf{N} & \textbf{Bloom distribution} & \textbf{Source seg.} & \textbf{Context chars} & \textbf{Rubrics} & \textbf{Prompt target} \\
\midrule
Payment Knowledge & 95 & L1: 95 & 1.0 & 1,180 & -- & Recall or recognize a source-specific payment rule in a self-contained four-choice MCQ. \\
Context-Grounded Reasoning & 96 & L2/L3/L4/L5: 38/29/19/10 & 1.9 & 3,116 & 3.1 & Explain, apply, analyze, or resolve exceptions/precedence over a supplied evidence pack. \\
Task-Input Attack & 115 & L2/L3/L4/L5: 18/88/6/3 & 1.9 & 3,593 & 3.2 & Apply one or two answer-preserving rule-applicability perturbations to a validated seed. \\
\bottomrule
\end{tabularx}
}
\end{table*}

\begin{table*}[h]
\section{Case Metadata Schema}
\centering
\caption{\textbf{Release metadata schema.} Required fields are included in the public benchmark package. Audit fields are released to support reproducibility and error analysis when they do not expose restricted source material or reviewer-identifying information.}
\label{tab:schema}
\footnotesize
\setlength{\tabcolsep}{4pt}
\renewcommand{\arraystretch}{1.06}
\begin{tabularx}{\textwidth}{@{}P{0.18\textwidth}P{0.16\textwidth}Y@{}}
\toprule
\textbf{Field group} & \textbf{Status} & \textbf{Fields} \\
\midrule
Identity and split & Required & Case ID, original construction ID when available, release split, benchmark subset, question type, Bloom level, and diagnostic category. \\
Evaluation input & Required & User-facing question or task input, source-grounded context for Open or Attacked Open cases, and choices for multiple-choice cases. \\
Answer and scoring & Required & Reference answer, grading rubrics for OpenQA, attack pattern codes for Attacked Open cases, and deterministic parsing rules for MCQ cases. \\
Source provenance and evidence & Required & Source-grounded evidence snippets used for evaluation, source category, source document ID, public URL when available, retrieval date, and evidence identifiers; full original documents are redistributed only when licensing permits. \\
Construction metadata & Audit release & Evidence-pack roles, scenario frame, chunk profile summaries, generation prompts, quality-gate outputs, and rejection reasons, with restricted source text or reviewer-identifying notes removed when needed. \\
Evaluation artifacts & Audit release & Evaluation prompts, model outputs, final judge decisions, manual audit labels, review notes when releasable, and run configuration; raw chain-of-thought is not stored or released. \\
\bottomrule
\end{tabularx}
\end{table*}

\begin{table*}[p]
\section{Detailed Attack Patterns}
\label{sec:attack_patterns}

\medskip
\centering
\small
\caption{\textbf{Detailed Attack Patterns for Task-Input Attack Item Generation.} The table lists both expert-initialized and pipeline-evolved attack modes, detailing their disruptive mechanisms and construction logic. \protect Mode E denotes Prior Inversion, whereas D\_E denotes Semantic Split + Prior Inversion; Figure~\ref{fig:attack_modes_heatmap} combines only the low-count D\_E/G display rows as D\_E/G (3/1).}
\label{tab:attack_patterns}

\setlength{\tabcolsep}{4pt}
\renewcommand{\arraystretch}{1.15}

\begin{tabularx}{\textwidth}{@{}c
>{\raggedright\arraybackslash}p{0.15\textwidth}
>{\raggedright\arraybackslash}p{0.29\textwidth}
>{\raggedright\arraybackslash}X
>{\raggedright\arraybackslash}p{0.07\textwidth}@{}}
\toprule
\textbf{Mode} & \textbf{Name} & \textbf{Mechanism} & \textbf{Construction} & \textbf{Source} \\
\midrule

A & Sub-rule Premise Collapse &
Global sub-rule S2 presupposes that parent rule S1 is universally valid. When the domain of S1 is narrowed, S2's premise vanishes, but the model still mechanically triggers S2. &
(1) Find a rule where a subset prioritizes a specific action;
(2) narrow the eligible pool to a single type so that S1 vanishes;
(3) retain triggers for S2;
(4) correct behavior: fallback to general sorting; error: applies S2 mechanically. &
Domain Experts \\

\midrule
B & Rule Coverage Gap Over-generalization &
Rule R covers general cases but remains silent on the current edge case E. The model logically extends R to fill the silence without realizing it is inventing a rule. &
(1) Find rule R lacking precise boundary constraints;
(2) place a distant rule R$'$ implicitly governing E;
(3) design scenario E to fall inside R's silent zone while appearing to be a general case. &
Domain Experts \\

\midrule
C & /Sub-section Override &
The main table provides base definitions, and the model stops reading once processed. Critical constraints are hidden in distant sub-sections. &
Choose values that trigger the hidden sub-section constraints. Ensure the sub-section is physically separated from the main table by other content. &
Domain Experts \\

\midrule
D\_E & Semantic Split + Prior Inversion &
Two segments share identical topic structures but differ in one qualifier, yielding opposite conclusions. The correct conclusion contradicts common sense. &
(1) Find rules R-alpha and R-beta with identical topics but opposite conclusions;
(2) design a scenario triggering R-beta;
(3) make R-alpha's wording prominent. &
Domain Experts \\

\midrule
E & Prior Inversion &
A knowledge-base rule conflicts with pre-trained common-sense expectations. The model finds the rule but silently overrides it. &
(1) Find a rule contradicting common sense;
(2) design a scenario where prior expectations conflict with the rule;
(3) emphasize ``according to the rule'' in the prompt. &
Domain Experts \\

\midrule
F & Symmetry Bias &
The correct answer requires treating superficially identical objects differently. The model tends to give a symmetric ``both can/cannot'' response. &
Design a situation where A and B belong to the same category, but A can perform an action while B cannot. The correct answer must treat them asymmetrically. &
Domain Experts \\

\midrule
G & Total vs.\ Batch / Sub-account &
A rule is triggered by insufficient sub-account or batch balances, but the model incorrectly uses the total balance for its judgment. &
Initialize an insufficient sub-account balance but sufficient total balance. Emphasize ``balance'' rather than ``batch balance'' in the prompt to induce confusion. &
Pipeline Evolution \\

\midrule
H & Phantom State Mutation &
An operation triggers an implicit global hook to switch the system into state S$'$. The model ignores S$'$ and processes subsequent steps using the original state. &
(1) Find an event switching the system to state S$'$;
(2) hide this event in intermediate steps;
(3) ask questions depending on state S$'$ rules, while hiding the trigger in sub-sections. &
Pipeline Evolution \\

\midrule
I & Causal Graph Hijacking &
The model has a strong causal prior A $\rightarrow$ C. System rules insert block B. The scenario bypasses B, meaning A cannot reach C. &
(1) Identify a strong causal prior;
(2) introduce M where X leads to Y only if M is passed;
(3) design a scenario bypassing M via obscure rules, thereby exploiting the model's faulty prior. &
Pipeline Evolution \\

\midrule
M & Polysemic Drift &
Identical business terms are assigned different precise meanings across documents. The model incorrectly mixes these contextual meanings. &
Find identical terms with varying definitions (e.g., the calculation of ``application date''). Construct a scenario at the boundary that requires contextual word-sense resolution. &
Pipeline Evolution \\

\bottomrule
\end{tabularx}
\end{table*}

\begin{table*}[!t]
\section{Evaluation Model Settings}
\label{app:model_settings}
\centering
\caption{\textbf{Model invocation settings for the final merged evaluation.} All answer-model calls use pass@1 and temperature 0. The final report keeps provider-default, explicit non-thinking, and supplemental thinking-enabled runs separate when multiple modes were evaluated.}
\label{tab:model_settings_appendix}
\scriptsize
\setlength{\tabcolsep}{3pt}
\renewcommand{\arraystretch}{1.05}
\begin{tabularx}{\textwidth}{@{}P{0.22\textwidth}P{0.13\textwidth}P{0.25\textwidth}P{0.15\textwidth}Y@{}}
\toprule
\textbf{Reported model or group} & \textbf{Platform} & \textbf{API model ID(s)} & \textbf{Mode in final report} & \textbf{Reasoning / storage note} \\
\midrule
GPT-5.5 & OpenRouter & \texttt{openai/gpt-5.5} & default/thinking enabled &  Provider usage reports hidden reasoning tokens for most rows. \\
Gemini 3.1 Pro Preview & OpenRouter & \texttt{google/gemini-3.1-pro-preview} &default/thinking enabled & Provider usage reports hidden reasoning tokens for a subset of rows. \\
Claude Opus 4.7 & OpenRouter & \texttt{anthropic/claude-opus-4.7} & thinking enabled & Enables reasoning; no raw chain-of-thought is stored. \\
\midrule
DeepSeek V4 Pro & Bailian & \texttt{deepseek-v4-pro} & default & Bailian default setting was used; older merged rows do not expose reasoning-token metadata. \\
Qwen3.5 family & Bailian & Qwen3.5 397B-A17B, 122B-A10B & default & Provider-default hybrid-thinking mode; usage metadata reports reasoning tokens when available. \\
GLM 5.1 & Bailian & \texttt{glm-5.1} & default & Provider-default mode; usage metadata reports reasoning tokens for a subset of rows. \\
MiniMax M2.5 & Bailian & \texttt{MiniMax-M2.5} & default & Provider-default mode; usage metadata reports reasoning tokens for a subset of rows. \\
\midrule
Qwen3 open-weight scale sweep & Bailian & Qwen3 4B, 8B, 14B, 32B, 235B-A22B & non-thinking; thinking enabled & Non-thinking rows use \texttt{enable\_thinking=false}; supplemental rows use explicit thinking. Both are reported for scale analysis. \\
Gemma 4 IT & OpenRouter & Gemma 4 31B IT; 26B-A4B IT & default / no explicit thinking; thinking enabled & Both rows are reported. Supplemental rows explicitly enable reasoning through OpenRouter. \\
Other overseas open models & OpenRouter & Llama 4 Maverick; Mistral Large 3; Ministral 3 14B/8B & default & No explicit reasoning parameter was sent; no reasoning-token metadata was observed for these exact model IDs. \\
\midrule
Judge & self-deployed & DeepSeek V4 Flash & final structured verdict only & OpenQA answers are judged by DeepSeek V4 Flash. MCQ uses deterministic option parsing. The judge stores final JSON decisions only, not reasoning traces. \\
Cross-family judge sensitivity & Self-deployed & Qwen3.5 397B-A17B; Qwen3.6-27B & non-thinking; temperature 0 & Calibrated rubric-first re-scoring of fixed outputs from three non-DeepSeek answer models; not used for headline scoring. \\
\bottomrule
\end{tabularx}
\end{table*}

\clearpage

\begin{table*}[t]
\section{Additional Experiment Results}
\label{app:exp_templates}
\centering
\caption{\textbf{Construction funnel and expert admission.} Counts are de-duplicated expert-review finalists, not raw iterative generation attempts. The Normal pool is lower assurance and excluded from Pro evaluation.}
\label{tab:construction_funnel}
\scriptsize
\setlength{\tabcolsep}{4pt}
\renewcommand{\arraystretch}{1.06}
\begin{tabular}{@{}lrrrrr@{}}
\toprule
\textbf{Stage} & \textbf{Finalists} & \textbf{Pro} & \textbf{Held out} & \textbf{Admit. \%} & \textbf{Normal} \\
\midrule
Base & 356 & 191 & 165 & 53.7 & 891 \\
Attack & 150 & 115 & 35 & 76.7 & 355 \\
\midrule
Total & 506 & 306 & 200 & 60.5 & 1,246 \\
\bottomrule
\end{tabular}
\end{table*}

\begin{table*}[t]
\centering
\caption{\protect \textbf{Primary exclusion reasons for the 200 held-out finalists.} Each item is assigned one primary reason from expert labels and written comments.}
\label{tab:app_exclusion_reasons}
\scriptsize
\setlength{\tabcolsep}{3.0pt}
\renewcommand{\arraystretch}{1.05}
{
\begin{tabularx}{\columnwidth}{@{}Xr@{}}
\toprule
\textbf{Primary reason} & \textbf{Share} \\
\midrule
Provider-specific or low-transferability rule & 42.5\% \\
Volatile or under-supported source grounding & 14.5\% \\
Answer--rubric mismatch or overly strict grading & 13.0\% \\
Ambiguous or underspecified task conditions & 12.5\% \\
Near-duplicate or coverage-balancing hold-out & 7.0\% \\
Business-realism or sensitivity concern & 5.5\% \\
Other borderline issue & 3.0\% \\
Weak diagnostic value & 2.0\% \\
\bottomrule
\end{tabularx}
}
\end{table*}


\begin{table*}[t]
\centering
\caption{\protect \textbf{Cross-family judge sensitivity on fixed OpenQA outputs.} All three judges are applied under the same calibrated rubric-first re-scoring protocol; this analysis does not alter headline scores. Each Qwen cell reports correct outputs, score shift relative to DeepSeek V4 Flash, and item-level agreement.}
\label{tab:secondary_judge_agreement}
\scriptsize
\setlength{\tabcolsep}{4pt}
\renewcommand{\arraystretch}{1.06}
{
\begin{tabularx}{\textwidth}{@{}P{0.20\textwidth}rP{0.16\textwidth}P{0.27\textwidth}Y@{}}
\toprule
\textbf{Answer model} & \textbf{N} & \textbf{DeepSeek correct} & \textbf{Qwen3.5-397B} & \textbf{Qwen3.6-27B} \\
\midrule
GLM 5.1 & 307 & 205/307 & 201/307; $-1.30$ pp; $94.14\%$ & 206/307; $+0.33$ pp; $91.21\%$ \\
Gemma 4 26B-A4B IT & 307 & 195/307 & 186/307; $-2.93$ pp; $93.81\%$ & 181/307; $-4.56$ pp; $93.49\%$ \\
Gemini 3.1 Pro Preview & 307 & 225/307 & 218/307; $-2.28$ pp; $91.21\%$ & 222/307; $-0.98$ pp; $91.21\%$ \\
\midrule
\textbf{Pooled} & \textbf{921} & \textbf{625/921} & \textbf{605/921; $-2.17$ pp; $93.05\%$} & \textbf{609/921; $-1.74$ pp; $91.97\%$} \\
\bottomrule
\end{tabularx}
}
\end{table*}

\begin{table*}[t]
\centering
\caption{\textbf{Supplemental thinking-mode results.} Values are accuracies on the final Pro benchmark. L1-C is L1 Closed accuracy; L2-O is L2--L5 Open accuracy; Attack-O is Attacked Open accuracy.}
\label{tab:app_scaling_thinking}
\scriptsize
\setlength{\tabcolsep}{4pt}
\renewcommand{\arraystretch}{1.05}
\begin{tabularx}{\textwidth}{@{}P{0.32\textwidth}P{0.18\textwidth}rrrY@{}}
\toprule
\textbf{Model} & \textbf{Mode} & \textbf{L1-C} & \textbf{L2-O} & \textbf{Attack-O} & \textbf{Note} \\
\midrule
\multicolumn{6}{@{}l}{\textit{Qwen3 thinking-mode size sweep}} \\
Qwen3 4B & thinking & 43.2 & 51.0 & 26.1 & Smallest model remains weakest under attacked inputs. \\
Qwen3 8B & thinking & 44.2 & 58.3 & 36.5 & Open evidence use improves, but robustness remains limited. \\
Qwen3 14B & thinking & 43.2 & 58.3 & 46.1 & Better attacked-input robustness than 8B. \\
Qwen3 32B & thinking & 48.4 & 69.8 & 42.6 & Highest Open accuracy in the Qwen3 thinking sweep. \\
Qwen3 235B-A22B & thinking & 55.8 & 61.5 & 47.0 & Best attacked-input accuracy in the Qwen3 thinking sweep. \\
\midrule
\multicolumn{6}{@{}l}{\textit{Other supplemental thinking-mode runs}} \\
Claude Opus 4.7 & thinking & 68.4 & 82.3 & 74.8 & Strong attacked-input robustness among supplemental runs. \\
Gemma 4 31B IT & thinking & 57.9 & 82.3 & 74.8 & Matches Claude Opus 4.7 on Open and Attack-O in this run. \\
Gemma 4 26B-A4B IT & thinking & 61.1 & 82.3 & 60.9 & Strong Open accuracy with weaker attacked-input robustness. \\
Kimi K2.6 & thinking & 66.3 & 90.6 & 73.0 & Highest L2--L5 Open accuracy in the supplemental thinking runs. \\
\bottomrule
\end{tabularx}
\end{table*}

\begin{table*}[t]
\centering
\caption{\textbf{Bootstrap uncertainty estimates.} Values are percentages, and SE denotes bootstrap standard error over benchmark cases. $\Delta$ is CGR-OB minus TIA-OB; because these are separate subsets, its SE is a descriptive independent-resampling estimate.}
\label{tab:app_bootstrap_se}
\scriptsize
\setlength{\tabcolsep}{2.7pt}
\renewcommand{\arraystretch}{1.04}
\begin{tabular}{@{}P{0.25\textwidth}*{10}{r}@{}}
\toprule
\textbf{Model} & \textbf{PK-CB} & \textbf{SE} & \textbf{CGR-CB} & \textbf{SE} & \textbf{CGR-OB} & \textbf{SE} & \textbf{TIA-OB} & \textbf{SE} & \textbf{$\Delta$} & \textbf{SE} \\
\midrule
Gemini 3.1 Pro Preview & 80.0 & 4.1 & 14.6 & 3.6 & 89.6 & 3.1 & 81.7 & 3.6 & 7.9 & 4.8 \\
GPT-5.5 & 77.9 & 4.3 & 36.5 & 4.9 & 82.3 & 3.9 & 80.0 & 3.7 & 2.3 & 5.4 \\
Claude Opus 4.7 & 68.4 & 4.8 & 25.0 & 4.4 & 82.3 & 3.9 & 74.8 & 4.0 & 7.5 & 5.6 \\
DeepSeek V4 Pro & 60.0 & 5.0 & 16.7 & 3.8 & 81.3 & 4.0 & 80.0 & 3.7 & 1.3 & 5.5 \\
Qwen3.5 397B-A17B & 62.1 & 5.0 & 14.6 & 3.6 & 84.4 & 3.7 & 72.2 & 4.2 & 12.2 & 5.6 \\
GLM 5.1 & 65.3 & 4.9 & 15.6 & 3.7 & 74.0 & 4.5 & 70.4 & 4.3 & 3.6 & 6.2 \\
Qwen3.5 122B-A10B & 59.0 & 5.0 & 9.4 & 3.0 & 84.4 & 3.7 & 70.4 & 4.3 & 14.0 & 5.6 \\
MiniMax M2.5 & 53.7 & 5.1 & 9.4 & 3.0 & 75.0 & 4.4 & 60.9 & 4.6 & 14.1 & 6.3 \\
Mistral Large 3 & 47.4 & 5.1 & 8.3 & 2.8 & 64.6 & 4.9 & 51.3 & 4.7 & 13.3 & 6.7 \\
Llama 4 Maverick & 54.7 & 5.1 & 8.3 & 2.8 & 68.8 & 4.7 & 46.1 & 4.6 & 22.7 & 6.6 \\
Gemma 4 31B IT & 57.9 & 5.1 & 9.4 & 3.0 & 82.3 & 3.9 & 74.8 & 4.0 & 7.5 & 5.6 \\
Gemma 4 26B-A4B IT & 61.1 & 5.0 & 7.3 & 2.7 & 82.3 & 3.9 & 60.9 & 4.6 & 21.4 & 6.0 \\
Qwen3 32B & 48.4 & 5.1 & 5.2 & 2.3 & 69.8 & 4.7 & 42.6 & 4.6 & 27.2 & 6.6 \\
Ministral 3 14B & 49.5 & 5.1 & 2.1 & 1.5 & 57.3 & 5.0 & 37.4 & 4.5 & 19.9 & 6.8 \\
Qwen3 8B & 44.2 & 5.1 & 1.0 & 1.0 & 58.3 & 5.0 & 36.5 & 4.5 & 21.8 & 6.7 \\
Ministral 3 8B & 51.6 & 5.1 & 2.1 & 1.5 & 50.0 & 5.1 & 29.6 & 4.3 & 20.4 & 6.6 \\
\bottomrule
\end{tabular}
\end{table*}

\clearpage
\begin{figure*}[!t]
\section{Closed/Open Diagnostic Details}
\centering
\includegraphics[width=0.98\textwidth]{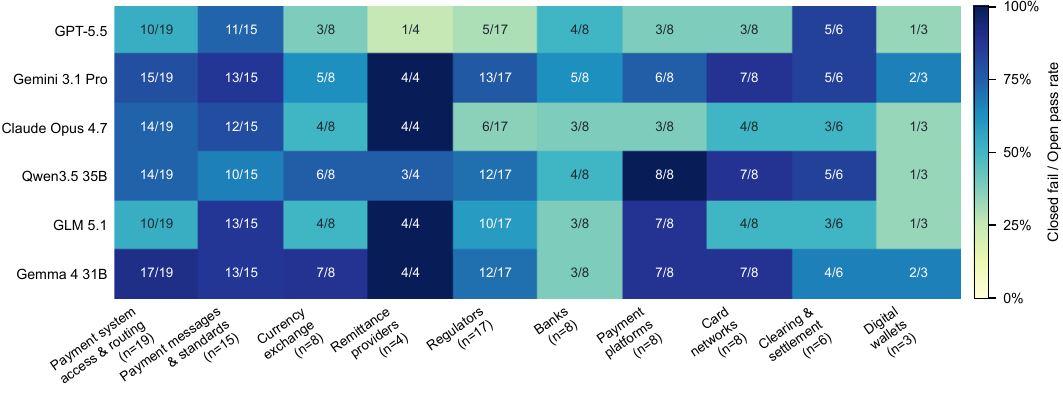}
\caption{\textbf{Where Open context rescues representative models.} Each cell counts L2--L5 cases where a model fails Closed but passes Open, grouped by the case's primary source category. Larger counts indicate source areas where the model lacks accessible parametric knowledge but can use the supplied source context.}
\label{fig:app_context_rescue_heatmap}
\end{figure*}

\begin{table*}[!t]
\centering
\caption{\textbf{Manual audit of representative Open failures.} Counts are model--case errors for six representative models on L2--L5 Open cases. The audit separates substantive context-use errors from residual judge/rubric sensitivity.}
\label{tab:app_open_failure_modes}
\scriptsize
\setlength{\tabcolsep}{4pt}
\renewcommand{\arraystretch}{1.08}
\begin{tabularx}{\textwidth}{@{}P{0.22\textwidth}ccP{0.31\textwidth}Y@{}}
\toprule
\textbf{Failure mode} & \textbf{Model-case} & \textbf{Cases} & \textbf{What failed} & \textbf{Typical cases} \\
\midrule
Multi-rule synthesis & 26 & 12 & Retrieves local facts but fails to combine rules, precedence, workflow order, or causal relations. & CPMI stability, CHAPS/CLS fallback, SVF onboarding \\
Scope or exception boundary & 26 & 10 & Misapplies an exclusion, exception, entity role, jurisdictional scope, or transaction-state boundary. & CHATS member exclusion, all-or-none, reporting boundary \\
Over-expanded answer path & 18 & 7 & Adds extra rules, exceptions, or operational actions that change or obscure the requested decision. & Visa OCT, Buna access, PayPal refund \\
Modal or threshold precision & 17 & 10 & Confuses may/must, possible/certain, immediate/conditional, or a numeric/date threshold. & Zelle possible delay, ISO 20022 may need, 21-day conversion \\
Judge/rubric sensitivity & 7 & 5 & The answer is substantially correct, but automatic scoring is sensitive to extra wording or missing exact phrasing. & offshore RMB conversion, same-day bill payment \\
Liability or responsibility boundary & 5 & 3 & Does not cleanly assign responsibility among platforms, banks, providers, or users. & Western Union wallet, BOCHK e-CNY, gateway vs escrow \\
\bottomrule
\end{tabularx}
\end{table*}

\begin{table*}[t]
\centering
\caption{\protect \textbf{Representative diagnostic cases.} Each row connects an observed model--case transition to the concrete payment rule that failed and the resulting triage implication.}
\label{tab:app_diagnostic_cases}
\scriptsize
\setlength{\tabcolsep}{3.3pt}
\renewcommand{\arraystretch}{1.08}
{
\begin{tabularx}{\textwidth}{@{}P{0.15\textwidth}P{0.18\textwidth}P{0.39\textwidth}Y@{}}
\toprule
\textbf{Signal} & \textbf{Model / case} & \textbf{Concrete failure} & \textbf{Triage implication} \\
\midrule
Context Rescued & Gemini 3.1 Pro / OTC Clear liability & In Closed, the model reverses the source rule by assigning responsibility to clearing members and exempting OTC Clear. With context, it correctly states that OTC Clear is responsible for item accuracy while SI, HKICL, and MA are exempt. & Supply the exact source rule when generic liability priors conflict with scheme-specific allocation. \\
Context-Use Gap & Gemini 3.1 Pro / HKICL all-or-none & The Open answer explains unwind and re-settlement, but incorrectly defines the general-member branch as excluding only HKSCC; the rule also excludes HKCC and SEOCH. & Verify entity scope and exception lists after retrieving the relevant rule. \\
Task-Input Attack & Qwen3 32B / MoneyGram debit timing & The task foregrounds a 4:30 p.m. submission and removes common delay factors. The source says the ACH cut-off does not apply to debit cards and gives no same-day guarantee, yet the model infers same-day debit and gives an internally inconsistent conclusion. & Check rule applicability before applying a familiar threshold or workflow prior. \\
\bottomrule
\end{tabularx}
}
\end{table*}

\clearpage
\section{Human Review Instructions}
\label{app:human_review_instructions}

Human reviewers assessed benchmark validity and judged correctness, not payment, legal, compliance, or customer-facing advice. Review artifacts included the question, supplied source context, reference answer, grading rubric, model output, judge verdict, and source identifiers when available. Reviewers marked whether an item was source-supported, answerable, unambiguous, payment-specific, diagnostically useful, and fairly scorable; Attacked Open review additionally checked that the perturbation preserved the source-supported answer path.

Four practitioners first jointly reviewed 30 calibration cases (10 per subset) to align application of this form. The remaining candidates were assigned in four disjoint subsets for one-practitioner first-pass review, then merged and jointly adjudicated. The form supported \textsc{Admit}, \textsc{Hold Out}, and \textsc{Revise}; an item was admitted only when adjudication left no unresolved issue. We define a disputed item operationally as one with an unresolved concern about source support, answer/rubric validity, ambiguity, diagnostic value, duplicate coverage, source stability, or attack validity after discussion. Such items were held out. Cases answered incorrectly by Gemini 3.1 Pro, GPT-5.5, or Claude Opus 4.7 received an additional two-expert validity review, and unresolved item-defect cases were removed.

For the sample-based proxy IAA, 60 candidates (20 per subset) were randomly sampled. Each was re-reviewed by an original payment-domain practitioner who had not conducted its first-pass review and was blind to the original label, using the same criteria. Binary admission agreement was 57/60 ($95.0\%$; Cohen's $\kappa=0.900$). This is a reliability check of the admission decision on a sample, not full-pool IAA.

Reviewers were informed that all materials were for benchmark construction and research evaluation only, that no private customer records, account data, transaction logs, or personally identifiable information should be used, and that benchmark cases should not be interpreted as operational certification or professional advice.

\end{document}